\documentclass[a4paper,conference]{IEEEtran}
\ifCLASSINFOpdf
   \usepackage[pdftex]{graphicx}
\else
\fi
\usepackage{amssymb}
\usepackage{amsmath}
\usepackage[ruled,vlined]{algorithm2e}
\usepackage{booktabs}
\ifCLASSOPTIONcompsoc
 \usepackage[caption=false,font=normalsize,labelfont=sf,textfont=sf]{subfig}
\else
 \usepackage[caption=false,font=footnotesize]{subfig}
\fi
\begin{document}
%
\title{Revisiting Optical Flow Estimation in 360 Videos}

\author{\IEEEauthorblockN{Keshav Bhandari,
Ziliang Zong,
Yan Yan}
\IEEEauthorblockA{Department of Computer Science, Texas State University, USA}}


\maketitle

\begin{abstract}
Nowadays 360 video analysis has become a significant research topic in the field since the appearance of high-quality and low-cost 360 wearable devices. In this paper, we propose a novel LiteFlowNet360 architecture for 360 videos optical flow estimation. We design LiteFlowNet360 as a domain adaptation framework from perspective video domain to 360 video domain. We adapt it from simple kernel transformation techniques inspired by Kernel Transformer Network (KTN) to cope with inherent distortion in 360 videos caused by the sphere-to-plane projection. First, we apply an incremental transformation of convolution layers in feature pyramid network and show that further transformation in inference and regularization layers are not important, hence reducing the network growth in terms of size and computation cost. Second, we refine the network by training with augmented data in a supervised manner. We perform data augmentation by projecting the images in a sphere and re-projecting to a plane. Third, we train LiteFlowNet360 in a self-supervised manner using target domain 360 videos. Experimental results show the promising results of 360 video optical flow estimation using the proposed novel architecture.
\end{abstract}


%
\IEEEpeerreviewmaketitle

\section{Introduction}
The immersive 360 video technology shows promising growth in the past years. Services such as GoPro, VeeR, Visbit, Facebook360 and YouTube have become great platforms for 360 videos. 360 videos are shaping the future of content creation and sharing. Hence, 360 videos will be an important digital medium in near future. This adds newer challenges and opportunities in computer vision research. One of such challenge is the motion and optical flow estimation in 360 videos.

Motion and optical flow estimation is important for 360 video understanding. Motion information can significantly aid tasks such as saliency detection, saliency prediction, gaze prediction, video piloting in 360 videos \cite{saliency1,saliency2, saliency6, saliency5}. Similarly, optical flow based panorama video stitching has shown impressive results compared with other methods. Deep Learning based optical flow estimation methods have shown significant improvement over classical methods \cite{horn, lucas}. 
Evolution of optical flow estimation methods from simple CNN based architecture to complex feature pyramid based architecture shows significant  improvements as well \cite{flownet, flownet2, liteflownet}.
\begin{figure}[t]
  \centering
  \includegraphics[width=0.5\textwidth]{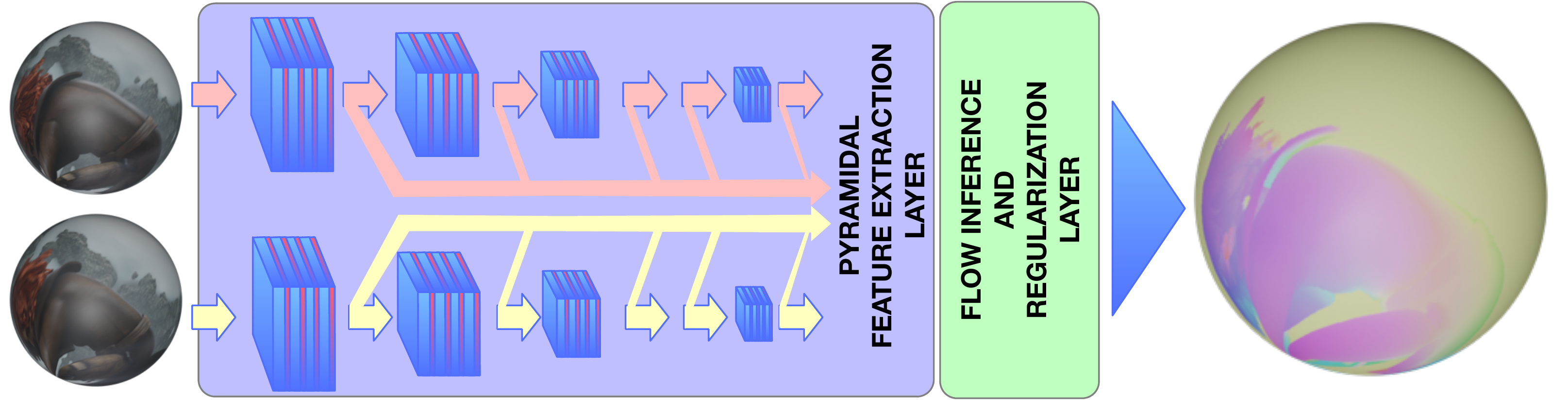}
  \caption{Simple representation of LiteFlowNet360 network architecture. We put focus only on feature extraction block which is shown in detail. Flow inference and regularization layer is similar to the original implementation. Input to the network are equirectangular or spherical data. Each convolution layer in pyramidal network is transformed to adapt spherical convolution(shown in red color). Final output is optical flow in spherical domain.}
  \label{arch1}
\end{figure}
However, regular CNN architectures are not suitable for 360 videos because of inherent distortion caused by projection of spherical videos to plane. We can use techniques such as \cite{sphconv_37, spectral_11, spectral_15, spherenet_46} to achieve spherical convolution. However, these methods have enormous overheads while converting existing complex pyramid based architecture to fit the needs of distortion free convolution for 360 videos. First, the training of the optical flow network is unstable. Having many transform convolution layers will lead entire process complicated as architectures becomes bigger. Second, we may not be able to guarantee that our model works even if we transform the architecture. We need some metrics such as EPE(End Point Error) to decide if our model works well. Since we lack labelled 360 video dataset for optical flow, the only method that fits our requirement is self-supervised methods. But how do we train this architecture in a self-supervised manner? The core part of self-supervised approach is calculating the loss between warped image and target image using predicted flow. Are warping techniques generalizable? In later sections we aim to answer these questions.

\begin{figure}[t]
  \centering
  \includegraphics[width = 0.5\textwidth]{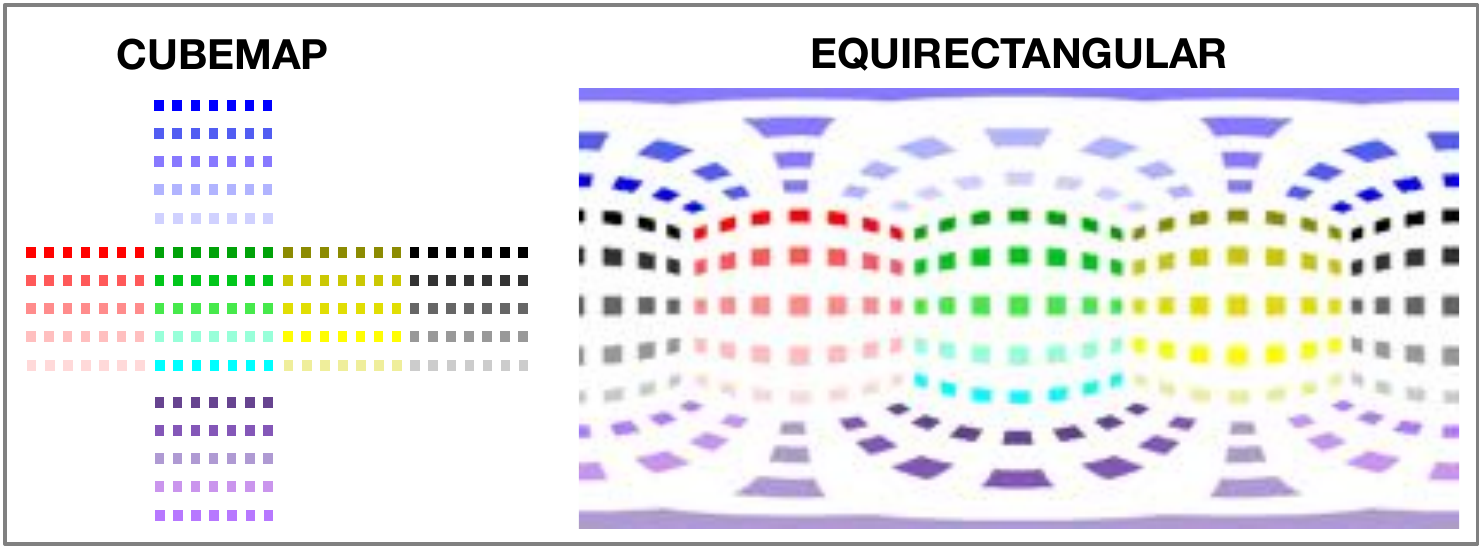}
  \caption{Showing how regular kernel map does not work in equirectangular (right) projection. When kernel applied in cubemap (left) are mapped into equirectangular projection it suffers huge distortions.}
  \label{kerneldistortion}
\end{figure}

Choosing a right architecture for our framework was the initial challenge we faced. However, we set certain requirements (like size, speed and efficiency) as a  guide to choose the right architecture. There are  many optical flow architectures to choose from, LiteFlowNet wins the competition. We will discuss more about this architecture in the following section. Framework we proposed would grow significantly as it includes significant changes as a part of perspective to a spherical domain transformation process. One more significant addition to this transformation process is the inclusion of special convolution to adapt the spherical nature for our dataset. This is important because the dataset we work on is an equirectangular plane, a sphere-to-plane projection. This planar projection incurs heavy distortion, which we have illustrated in Fig.\ref{kerneldistortion}. These special convolution, termed as spherical convolution, are expensive in terms of computation, which voids our requirements. Therefore, we adopt techniques like kernel transformation using transformer network. One such architecture \cite{ktn} dubbed as KTN has shown comparable improvement over later methods with less computation. We adopt KTN as our transformer network to learn spherical convolution.

Training an end-to-end optical flow architecture requires significant considerations of extra jobs like scheduling of training, implementing stacked architecture, considerations of motion magnitude and many other details to make it work on a par with the state-of-the-art results. Training architectures like this using better strategy to cope with gradually increasing task is an adoption of a popular philosophy, curriculum learning\cite{curr}. We have seen architecture like \cite{flownet2, liteflownet} adopted this strategy to create a better model. Apart from traditional optical flow estimation strategies, we have additional requirements because of the spherical nature of the dataset. Therefore, end-to-end training of optical flow for 360 video is highly unstable as there are many parallel objectives to achieve. We need to make sure that our model size does not grow significantly. This can slow down the training and inference speed. We need to address the nature of optical flow in 360 domain which can change the interpretation of warping techniques, flow representation and many other aspects. To make it brief, training optical flow architecture in 360 videos is not straightforward. To achieve the stable training process and fulfill our requirements, we adopt a divide-and-conquer strategy, thus dividing the entire training process into three major stages.

First stage of LiteFlowNet360 is to train LiteFlowNet architecture in perspective video datasets like \cite{sintel}. Then, we transform source CNN to target CNN layer wise. This transformation technique is progressive which means we need to do everything in-order. We will explain details about process of transformation in method section. After transforming into target CNN we will now further train entire model in an end-to-end fashion in supervised manner. To achieve this task we augment both source perspective videos and target optical flows into spherical distortion setting. When the second stage is complete, we will use a self-supervised scheme to further train our model in target videos. To do this, we need to perform a task like back-warping of frames using predicted flow. We use these predicted or warped frames as the basis of the training process by minimizing the similarity loss between ground frames and predicted frames. We adopt occlusion aware scheme inspired by \cite{selfflow1}. 

In this paper we exploit the existing optical flow estimation techniques and distortion free convolution in 360 videos. Our contributions are three folds: \textbf{(i)} To the best of our knowledge, this is the first work to address deep learning based dense optical flow estimation in 360 videos. \textbf{(ii)} We present an algorithm inspired by \cite{ktn} to transform learned representations from pre-trained network. \textbf{(iii)} We present a self-supervised learning approach since we do not have ground truth optical flow for 360 videos.

\section{Related Work}
\textbf{Optical Flow Estimation.}
The classical optical flow estimation approaches \cite{horn, lucas} used variational approaches to minimize energy based on brightness constancy and spatial smoothness. Recently,  \cite{flownet} proposed an end-to-end optical flow estimation with convolutional networks (FlowNet) using supervised scheme. Several other works based on CNN followed FlownNet including 3D convolution based approach \cite{end2end}, unsupervised approach \cite{unflow1,unflow2,unflow3} and pyramidal-coarse-to-fine approach \cite{pyra1,pyra2}. Recent variants such as \cite{patch1,patch2} used sparse matching by learning feature embedding. These methods were computationally expensive, making it impossible to train end-to-end fashion. FlowNet-2.0 \cite{flownet2} was an important addition in this series. It exploited curriculum learning approach \cite{curr}. In their work they address the weakness of FlowNet by addressing a smaller to larger range of displacement magnitude. However, the success of FlowNet-2.0 comes with a cost of over parametrization with around 160 Million parameters.  \cite{liteflownet} presented a more effective approach dubbed as LiteFlowNet. LiteFlowNet is around 30 times smaller and around 1.36 times faster. LiteFlowNet excelled FlowNet-2.0 by drilling down architecture details. LiteFlowNet proposed effective flow inference at each pyramid level, presented data fidelity and regularization as variational methods, whereas FlowNet only used a U-Net like architecture only. Self-supervised \cite{selfflow1, selfflow2} approaches for optical flow estimation are intuitive and reasonable approaches, as warping is one of the fundamental techniques used in successful deep learning based architecture. These techniques motivate our self-supervised learning scheme in the final stage.

\begin{figure*}[!t]
  \centering
  \includegraphics[width = 0.8\textwidth]{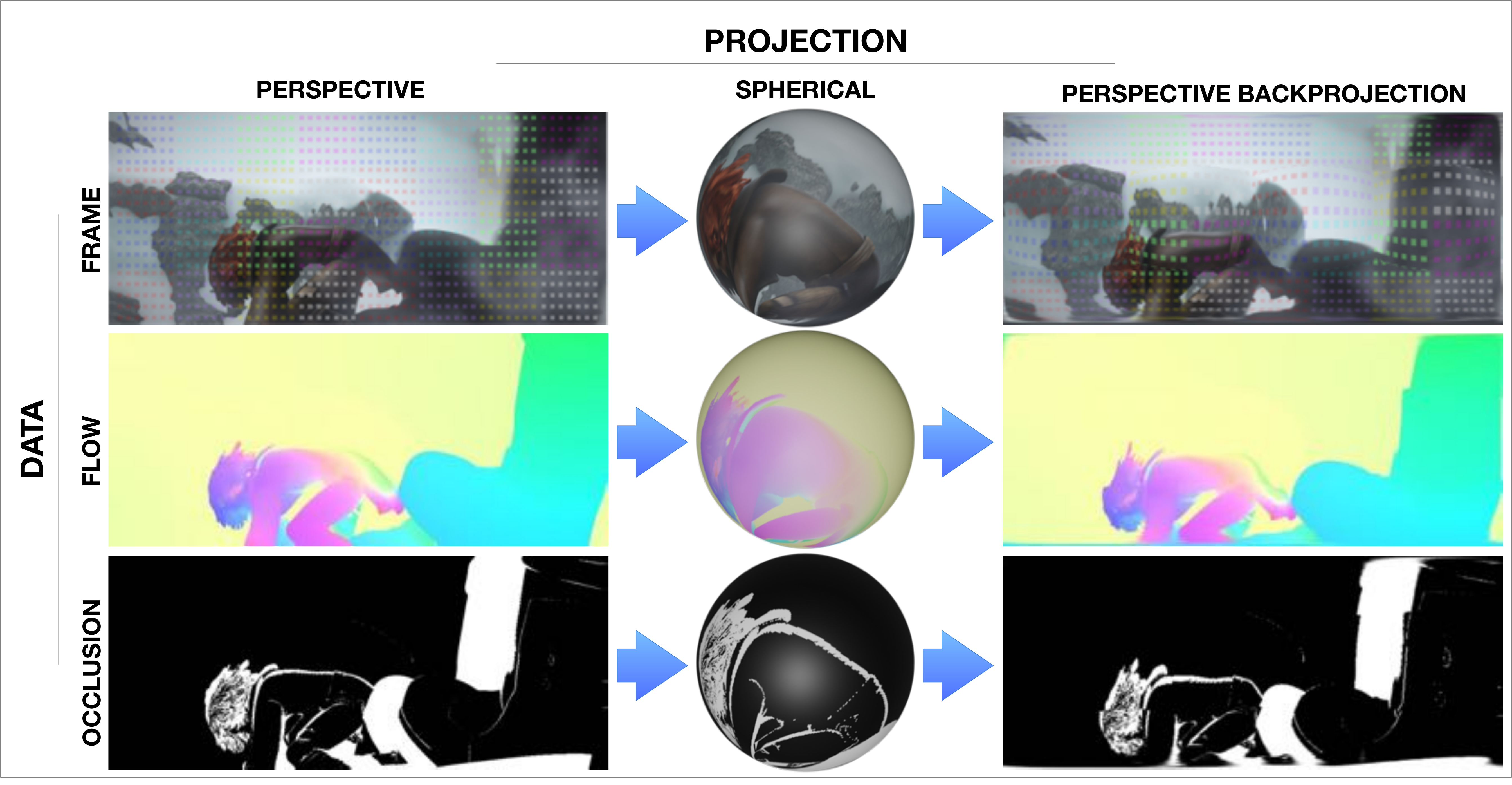}
  \caption{Spherical Data Augmentation. Perspective videos are projected in a unit sphere and then back projected to equirectangular plane. This is intentionally lossy process to create distortion artifact(shown in top row) in perspective data.}
  \label{sampledata}
\end{figure*}

\textbf{CNN for 360 Video/Images.} Performing direct convolution on spherical data led to inaccurate models \cite{pilot_19, hyper_30}. An intuitive approach to perform convolution on spherical data is to use convolution directly in cube map projection \cite{cube_3, cube_7}. This introduced less distortion but the model will have discontinuities, which led to sub-optimal model for several tasks. Another approach to learn rotation invariant CNN was to use graph convolution \cite{graph_24} techniques. This can be done by defining convolution in spectral domain \cite{spectral_11,spectral_15}. Similarly, this can also be done by projecting both feature maps and kernel in a spectral domain and apply regular CNN. These methods lose semantic information and were not useful in our case. In a recent year, several other spherical CNN based models have been proposed. Work such as \cite{ sphconv_37, spherenet_46} considered distortion in sphere-to-plane projection of spherical images/videos. Recent method by \cite{ktn}, dubbed as Kernel Transformer Network(KTN) is a significant piece of work in this domain. This architecture efficiently transferred convolution kernels from perspective images to the equirectangular projection. Basically, KTN produced a function parametrised by a polar angle and kernel as output. This work preserved the source CNNs and maintained accuracy, meanwhile offering transferability and scalability. Our architecture is a modified version of KTN, which uses interleaving convolution techniques to reduce discontinuity during convolution.

\textbf{360 Flow Estimation.}  \cite{omni1} proposed an approach by back-projecting image points to a virtual curved retina intrinsic to the geometry of central panoramic camera. Their method could adopt to contemporary ego-motion algorithms.   \cite{omni2} implemented Lucas-Kanade based method for optical flow estimation in catadioptric images. They proposed new constraint based on motion model defined on perspective images. This new constraint-based model was used to compute optical flow for omnidirectional image sequences. \cite{omni3} used multichannel spherical image decomposition techniques to compute optical flow for 360 image sequences. Similarly  \cite{omni5} proposed several variational regularization methods to estimate and decompose motion fields on the sphere.   \cite{omni4} implemented, adapted phase based method to compute optical flow using different treatments to account for 360 images. 

Our work fall somewhere in the intersection of optical flow estimation and emerging domain of omnidirectional computer vision. However, none of the above methods address the optical flow estimation problem using deep learning methods.

\section{Method}
We choose LiteFlowNet~\cite{liteflownet} as the basis for our newly proposed LiteFlowNet360 (shown in Fig.~\ref{arch1}) architecture because of its simpler design, lightweight (5.37M parameters) and highly efficient implementation. We represent our LiteFlowNet360 architecture in terms of two important blocks, feature extractor block ($F$) and regularization block ($P$) as shown in Eq.~\ref{featureblock} where $X$ is a sequence of two consecutive frames, $k$ is the number of layer transform from $0^{th}$ to $k^{th}$ layer in $F$ and where $n$ is an optimum number of layers eligible for transformation.

\begin{equation}\label{featureblock}
F_k(X) = 
     \begin{cases}
       F_0(X) & :k=0\\
       F_k(F_{k-1})(X) & :n > k>0\\
     \end{cases}
\end{equation}

Feature extractor block $F$ is transformed to adapt our need of 360 flow estimation as shown in Eq.~\ref{featureblockprime}. Each convolution layers $F_0,F_1, ..., F_{(n-1)}$ in feature extractor block is parametrised by a function $\Omega = g(\theta,\phi)$ in sphere, by generating different kernels for distortion above and below the equatorial region in sphere such that layers $F'_0,F'_1, ..., F'_{(n-1)}$ are our target layers. We compute location dependent kernel using polar and azimuthal angle $\theta$ and $\phi$ respectively.

We keep inference and regularization block as the same as LiteFlowNet. However, feature warping techniques at each pyramid level is transformed to address warping in 360 video domain. Further details will be explained in the following section.
\begin{equation}\label{featureblockprime}
F_k^\prime(X) = 
     \begin{cases}
       F_0^\prime(F_0(X),\Omega) & :k=0\\
       F_k^\prime(F_k,\Omega) (F_{k-1}^\prime(F_{k-1},\Omega))(X) & :n > k>0\\
     \end{cases}
\end{equation}

LiteFlowNet360 framework is an evolutionary architecture. We start from regular LiteFlowNet architecture and perform incremental transformation and training process to achieve final architecture. We formulate three important subsequent stages, transformation stage, intermediate refinement stage and final refinement.

\subsection{Stage 1: Transformation}
This stage starts with training the LiteFlowNet architecture with labeled data following  \cite{liteflownet}. The most important part of this stage is to transform convolution layers trained on perspective images to adapt to 360 images. We follow \cite{ktn} with some improvements, which we will present later in this section. Since we use equirectangular projection method, distortion depends only on polar angle. This leads to direct correspondence of the polar angle to the height of the input image, i.e., $y=\theta h / \pi$. This means we can utilize a single kernel for optimal row-group size($n_g$) such that we have $h/n_g$ projection matrices $P_i \in \rm I\!R^{r_i \times k_h \times k_w} $, where $r_i = h_i \times h_w$ is target kernel for each row-group $i$ and $(k_h\times k_w)$ is an original kernel size from source CNN as in \cite{ktn}. Different from original implementation, we interleave these rows as shown in Algorithm-\ref{interleave} to maintain connectedness, where $n_l$ is interleaving factor. We choose $n_l=3$ in our case.
\begin{equation}\label{loss}
\begin{array}{l}
     Y_k = F_k(X), 
     Y_k^\prime = F_k^\prime(F_k(X),\Omega)\\
    L_k = 
    ||Y_k^\prime - Y_k||^2\\
\end{array}
\end{equation}

\begin{equation}\label{lossperrowgroup}
    L_k' = \frac{1}{n_g}\sum_{i}^{n_g}{L_k(\Omega(Y_k'^i), Y_k^i)}
\end{equation}

We train each layer of feature extractor evolutionarily. We feed augmented images created by warping perspective image sequence as inputs to transformed layer and original image sequence as inputs to source CNN. Warping process is done by plane-to-sphere and sphere-to-plane projection of perspective image sequences. This back projection trechnique introduce distortion in perspective videos, as shown in Fig.\ref{sampledata} (top row). We train each layer with objective function presented in Eq.\ref{loss} to minimize the L2 norm between feature map generated by source CNN and transformed CNN layers, where $Y_k$ and $Y'_k$ represent output at $k_{th}$ layers in source and target architecture respectively and $L_k$ is an L2 norm between feature map from source and target CNN.

We project feature map row-group wise in tangential plane and compute loss with respect to corresponding feature map row-group from perspective source CNN. We combine these losses by averaging all the losses in row-group as shown in Eq.\ref{lossperrowgroup}, where $i$ refers to $i$-th row-group, and $L_k'$ is an L2 norm averaged over row-group.

\subsection{Stage 2: Intermediate Refinement}
\begin{figure}[t]
    \centering
    \includegraphics[width = 0.35\textwidth]{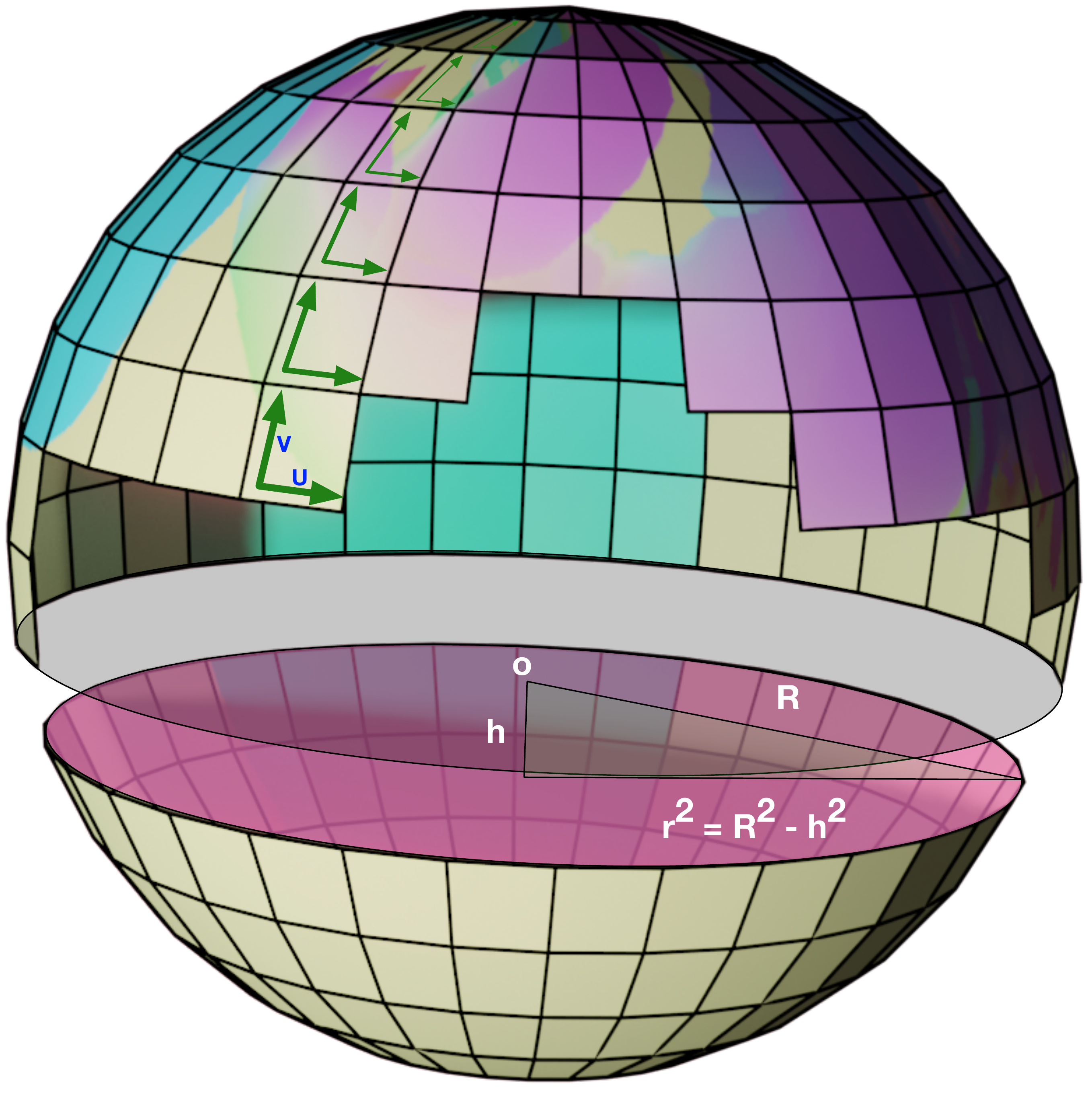}
  \caption{Flow representation in spherical domain. $(u,v)$ component changes as we move away from equator.}
  \label{sphere}
\end{figure}

Though first stage can transform convolution layers to adapt spherical images, it does not guarantee that estimated flow are well represented. A common problem will arise when we try to warp inferred flow around the sphere. This is due to the nature of sphercial coordinates. We can observe in Fig.~\ref{sphere} that the size and the shape of the patches decreases as we move away from the equatorial region. This means $u$ and $v$ component(shown in figure) changes as we move away from the equatorial region. This is because of the difference between idea behind the optical flow representation in perspective domain vs spherical domain. Optical flow in perspective domain is represented by the displacement in terms of euclidean distance. However, in spherical domain flow information makes sense only if we represent flow in terms of angular displacement. This means we need to present $u$ and $v$ in terms of $u_\theta$ and $v_\phi$ component. Instead of obtaining a direct solution, which is beyond the scope of our work, we introduce some correction factor on original $u$ and $v$ and project it in spherical domain.

\begin{algorithm}\label{interleave}
\SetAlgoLined
$Y\gets tensor()$\;
$X\gets input$\;
$tied\_weights\gets n_g$\;
$n\_transform\gets h/n_g$\;
\For{each $row \in [0, n\_transform]$ }{
    $start\gets row \times tied\_weights$\;
    \eIf{$row < (n\_transform - 1$)}{
        $end\gets start + tied\_weights+n_l$;
    }{
        $end\gets start + tied\_weights$\;
    }
    $Y[start:end]\gets \sum_{i,j}{K_{row}[i,j] * X_{row}[x-i,y-j]}$;
}
 \caption{Interleaving Convolution}
\end{algorithm}

\begin{figure*}[h]
    \centering
    \includegraphics[width = 0.75\textwidth]{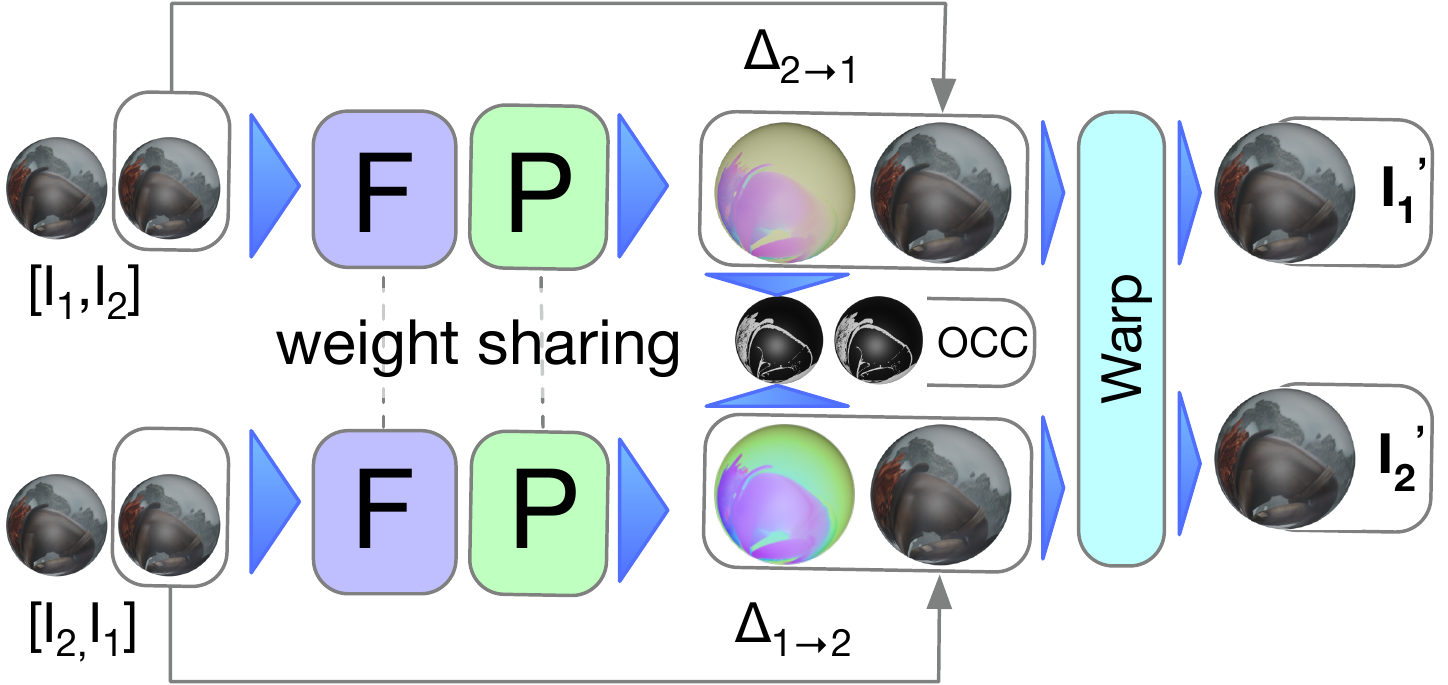}
  \caption{Final Refinement process. Network from second stage is extended to have two parallel weight sharing architecture.}
  \label{arch2}
\end{figure*}

\begin{algorithm}[h]\label{algoprojection}
\SetAlgoLined
$\Delta_{I_1\rightarrow I_2} \gets input()$\;
$(I_1,I_2)\gets I \gets input()$\;
$(h,w)\gets dim(I_1)$\;
$(r_w,r_h)\gets (\frac{h}{4\pi},\frac{w}{2\pi})$\;
\For{each $i \in [-h/2, h/2]$ }{
    $\Delta_{u} = \Delta_{I_1\rightarrow I_2}[i,:]\times \frac{2\pi \sqrt{r_w^2 - |r_w-i|^2}}{w}$\;
}
\For{each $j \in [-w/2, w/2]$ }{
    $\Delta_{v} = \Delta_{I_1\rightarrow I_2}[:,j] \times \frac{2\pi \sqrt{r_h^2 - |r_h-j|^2}}{h}$\;
}
$\Delta'_{I_1\rightarrow I_2} \gets \omega(\Omega(\Delta))$\;
$I'\gets(I'_1,I'_2) \gets \omega(\Omega(I))$\;
 \caption{Spherical Data Augmentation}
\end{algorithm}

The second stage is to refine the representation learning of optical flow in spherical domain. The intermediate refinement process is all about end-to-end training of the transform network. The training process is supervised as we want to make sure our network learn the actual representation. The problem with this scheme is that we do not have labelled dataset. Core part of the intermediate refinement stage is to use data augmentation techniques to convert labelled data, both images and optical flow in a spherical domain. We show sample augmented image sequences and corresponding optical flow in Fig.\ref{sampledata}.

Equirectangular plane is expressed from $(-\pi,\pi)$, $(-\pi/2,\pi/2)$ for length and height respectively, leading the aspect ratio of $length:height = 2:1$. We resize our original image and optical flow with the nearest interpolation scheme to maintain required aspect ratio. Then, we use simple projection techniques given by ($\phi = 2\times\pi\times u, \cos{\theta} = 2\times v - 1$) for unit sphere to perform forward projection (i.e., perspective to spherical projection) followed by restoration using backward projection (i.e., spherical projection to backward projection).

As we discussed, issues regarding projecting perspective optical flow directly into sphere requires a correction factor. We apply this correction factor separately for $u$ and $v$ component of original perspective flow. The idea behind these factors is to scale displacement magnitude to be fair all over the points in spherical representation. For example, $u$ is corrected by scaling each row with the ratio of central circumference (corresponding to the actual width $w$ in perspective plane) and circumference $w_i = 2\pi r_i$ at each row. Regarding calculation of $r_i$, see Fig.\ref{sphere}, where radius of a pixel-row $i$ at distance $h_i=\left|R-i\right|$ from center can be calculated using simple law of triangle $r_i^2 = R^2 -h_i^2$ where $R = \frac{w}{2\pi}$, where $i \in (-R,R)$. We finally define function $\Omega_{(x,y)}$ to perform perspective to spherical projection, $\omega_{(r,\theta,\phi)}$ to perform back projection and $\zeta$ as correction function. Now we present image augmentation $I$ as $I'=\omega(\Omega(I))$ and optical flow $\Delta$ as $\Delta'=\omega(\Omega(\zeta(\Delta))$. We present spherical data augmentation algorithm in Algorithm-\ref{algoprojection}.

The training objective is to minimize end point error $||\Delta'- \Delta''||$ between predicted flow $\Delta''$ and ground truth flow $\Delta'$ in conjunction with brightness error $||(I'_1+\Delta'_{I_1\rightarrow I_2})-I''_2||$ between the warped image and source image. We follow routine prescribed by \cite{liteflownet} to train our network, but we limit our training process to significantly fewer amount of epochs compared to original implementation, as this is only a refinement process and network plateaus in terms of error rate. Our model is now ready to cope with the spherical domain, but we need to adapt our model to real-world data. To adapt our model to real-world data, we move into ultimate refinement stage.
\subsection{Stage 3: Final Refinement}
\begin{figure*}[!t]
  \centering
  \includegraphics[width=\textwidth]{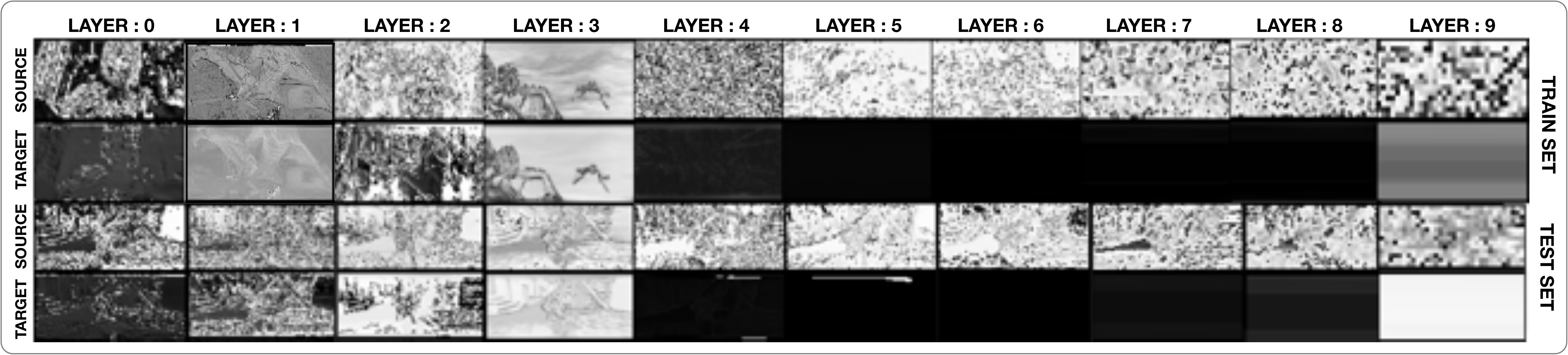}
  \caption{Showing randomly picked output individual channel from output of different layers in source and target(stage 2) architectures. First two rows represent source source and transform CNN respectively on train set, similarly second two rows represent test set.  }
  \label{activations}
\end{figure*}

\begin{algorithm}[t]\label{equiwarp}
\SetAlgoLined
$(g_\theta,g_\phi) \gets ([-180,+180], [-90,+90])$\;
$G \gets mesh\_grid(g_\theta, g_\phi)$\;
$(\tilde{\Delta}_u,\tilde{\Delta}_v)\gets \tilde{\Delta} \gets G + \Delta_{1\rightarrow 2}$\;
$\tilde{\delta_u} = \frac{\tilde{\Delta_{u}}}{|\tilde{\Delta_{u}}|}\small(|\tilde{\Delta_{u}}|-360)$\;
$\tilde{\delta_v} = \frac{\tilde{\Delta_{v}}}{|\tilde{\Delta_{v}}|}(180 - |\tilde{\Delta_{v}}|)$\;

$\tilde{\Delta}_{u} = 
\begin{cases}
`\tilde{\Delta}_{u}, & \tilde{\Delta_u} \in [-180,180]\\
-\tilde{\Delta}_{u}, & \tilde{\Delta_v} \notin [-90,90]\\
\tilde{\delta_u}, &  \tilde{\Delta_u} \notin [-180,180]\\
\end{cases}$\;

$\tilde{\Delta}_{v} = 
\begin{cases}
\tilde{\Delta}_{v}, & \tilde{\Delta_v} \in [-90,90]\\
\tilde{\delta}_{v}, & \tilde{\Delta_v} \notin [-90,90]\\
\end{cases}$\;
$\tilde{\Delta} \gets (\tilde{\Delta}_u,\tilde{\Delta}_v)$\;
 \caption{Boundary Condition}
\end{algorithm}
We replicate our initial network from stage-2 into two channel siamese network as shown in Fig.~\ref{arch2} to estimate forward $\Delta_{1\rightarrow 2}$ and backward $\Delta_{2\rightarrow 1}$ flow. We use forward and backward flow to estimate occlusion $\tilde{O} = (\tilde{O}_{2\rightarrow 1}, \tilde{O}_{1\rightarrow 2})$ using Eq.\ref{occlusioneq} where $\epsilon \approx 10^{-2}$, $(i,j) = (1,2)$ for forward flow and vice versa. 
\begin{equation}\label{occlusioneq}
    \begin{array}{l}
        M_i = \begin{cases}
        0 \\
        1 & if, |\Delta_{i\rightarrow j}| \leq \epsilon\\
        \end{cases}\\
        \tilde{O}_{i\rightarrow j} = M_i \odot ((1 - M_j)+\tilde{O}_{j\rightarrow i})\\
        \\
        L_p = \sum_{i,j}{\frac{\sum{\psi(I_i - I'_i)}\odot(1-O_{i\rightarrow j})}{\sum{1-O_{i\rightarrow j}}}}\\
    \end{array}
\end{equation}
Similarly, we use predicted optical flow to warp target image. Apart from traditional warping techniques, we modify warping technique as shown in Algorithm-\ref{equiwarp}. This warping technique is necessary to address the continuous nature of 360 images, i.e., whenever pixel displacement occurs beyond the boundary condition the pixel is displaced somewhere within the equirectangular plane. For example, if a pixel is displaced beyond the right boundary, the pixel will be displaced on the left side of the equirectangular plane. This is not true with perspective flow, where we consider this as a boundary condition and put the pixel into boundary. This is well preserved and more accurate assumption for smaller displacement in the border area.

We present final refinement process as further training steps to adapt to the target domain. We use dataset from our ongoing work Egok360, an egocentric activity recognition dataset for 360 videos as target dataset. The training process is self-supervised based on photometric loss as shown in Eq.\ref{occlusioneq} where $\psi = (|x|+\epsilon)^q, \epsilon \approx 10^{-2}, q \approx 1\times 10^{-1}$.

\section{Results}
We present our result mainly on augmented Sintel~\cite{sintel} dataset, which we termed as Sintel360. We performed spherical data augmentation on original Sintel training set, which we divided into 9:1 train-val set. This train-val set has ground truth optical flow information. We compared 4 different models as shown in Table~\ref{quantitative} using commonly used end point error (EPE) metrics using validation set. To make comparision fair, we augmented original sintel test set. We used this test set to compute photometric loss ($L_p$), defined in Eq.\ref{occlusioneq}. 
\begin{table}
\caption{Experimental results on Sintel360 dataset.}
  \label{quantitative}
  \centering
  \begin{tabular}{lllll}
    \toprule
    Model     & Data     & \#Layers & $EPE$ & $L_p^*$\\
    \midrule
    LiteFlowNet\cite{liteflownet} & Sintel360  &  $0$ & $\sim$ 6.35 & $\sim$ 1.30\\
    LiteFlowNet +\cite{ktn} & Sintel360     & $>4$  & $\geq$ 17 & $\geq$ 3.06\\
    \textbf{Ours}, Stage-2 & Sintel360    & $4$   & $\sim$ 6.35 & $\sim$ 0.70\\
    \textbf{Ours}, Final     & Sintel360 & $4$ & $\sim$ 3.95 & $\sim$ 0.60    \\
    \bottomrule
  \end{tabular}
\end{table}

\textbf{Quantitative Results.} Table~\ref{quantitative} summarizes our experiments. We found that exhaustive layer replacement task is unnecessary. The convergance rate dramatically decreases as we go deeper, as shown in Table~\ref{quantitative}, EPE is significanly higher ($\geq17$) for more than $4$ layers replacement. This creates a domino effect, which propagate errors in subsequent layers. We illustrate this effect in Fig.\ref{activations}. We can see that beyond layer 4, the output of transformed layers are different. Instead of reproducing the source CNN these layers learn nothing even after training for 30 epochs, using same techniques that was used to train previous layers.

Our model on Stage 2 performs on par with original implementation. Though original method seems fine, representation for optical flow in spherical video is not fair. We can explain the lower EPE on the original model with the large number of flow information correspondence between real and augmented data in central region of equirectangular plane. With a final refinement stage, we improve our model significantly bringing EPE to 3.95 from 6.35 on val-set and photometric loss from 0.70 to 0.60 on a test set.
\begin{figure*}[!t]
  \centering
  \includegraphics[width=0.8\textwidth]{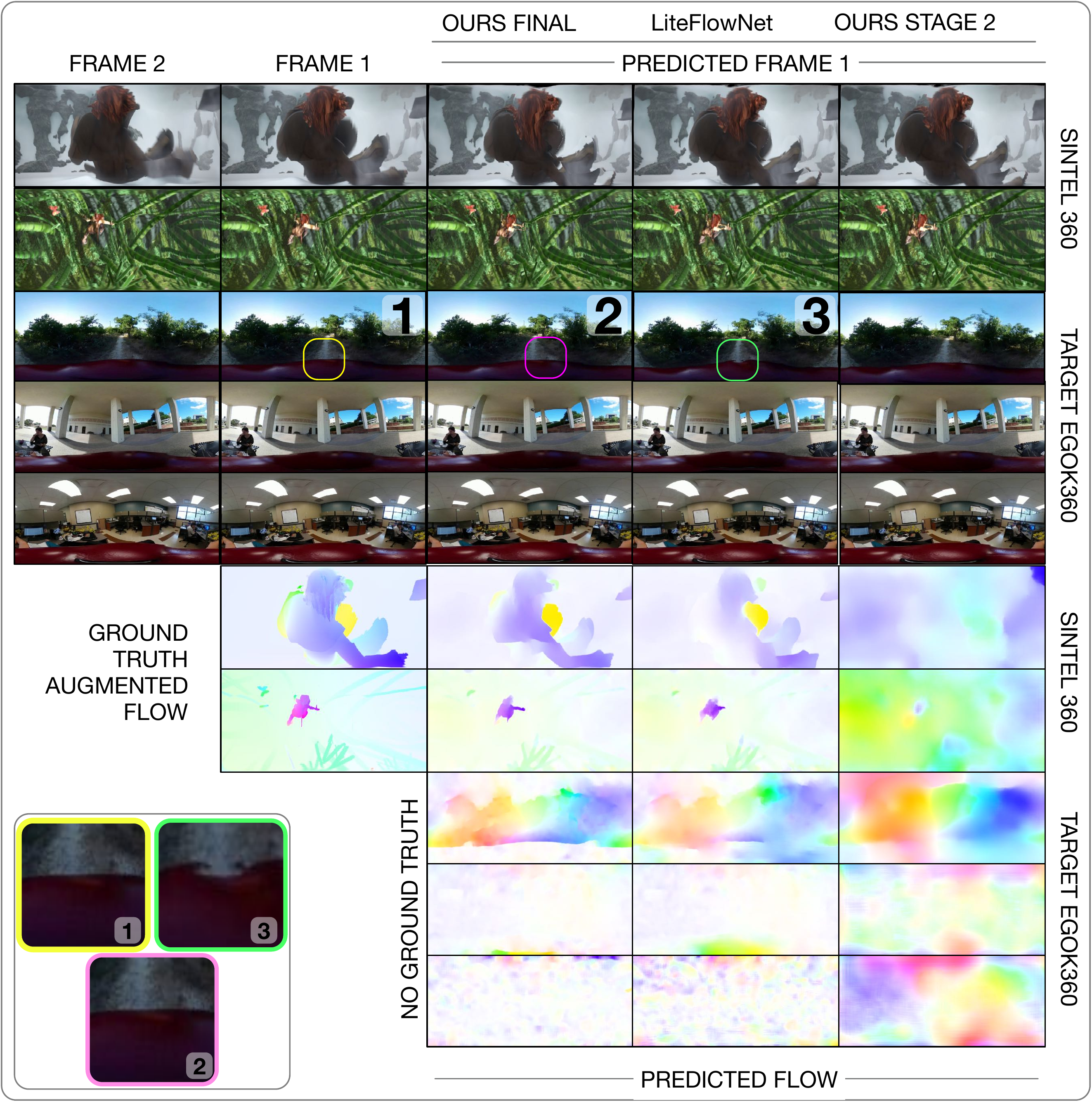}
  \caption{Qualitative results on augmented Sintel 360 dataset and target video dataset. First two row represents randomly picked frames and second two row represents corresponding optical flow information. We predict frame-1 using forward flow from each architecture. We randomly pick patches from same location from predicted(patch-2,patch-3) and ground truth(patch-1) frame-1 as shown in bottom left corner. Patch 2,3 are from liteflownet360 and liteflownet respectively. We can see liteflownet360 results are comparatively better. Note: We encourage digital reader to zoom in for detail view.}
  \label{qualitative}
\end{figure*}

\textbf{Qualitative Result.} Fig.\ref{qualitative} shows qualitative results from our experiment compared with baseline LiteFlowNet. We also present qualitative results on our target 360 video dataset. 
To understand the fairness of the flow predicted, we used flow information to predict the next frame. We observe that the warping of flow preserves the spherical nature. In another word, it preserves the artificial artificat we introduced in original dataset (please note patches in different colors in target dataset shown in Fig.\ref{qualitative}). However, there are cases where none of these models work as expected. We show such case in the last row of estimated optical flow on target dataset. We believe this can be improved further by allowing the model to have longer training times with further hyperparameter exploration.

\section{Conclusion}
In this paper, we presented a novel framework for 360 optical flow estimation, dubbed as LiteFlowNet360. This framework is an adaptation of existing best practices from both of the world, ``optical flow estimation for perspective videos'' and ``spherical convolution for 360 videos/images''. We presented our work as three major subsequent stages, transformation stage, intermediate refinement stage and final refinement stage. We started with the process of transformation, which includes evolutionary learning of spherical convolution based on transformer network. Apart from the success of these methods in other field, we empirically showed that exhaustive layer transformation from source to target CNN is insignificant in the context of optical flow estimation. We present second stage to address the correct representation of 360 flow. This stage requires further training as a refinement task. To train our model, we introduce a lossy data augmentation techniques to exploit existing labelled datasets. This technique allowed us to introduce artifacts related to spherical distortion in perspective videos, meanwhile transforming optical flow information in a spherical domain. We presented final stage as a domain transfer stage, where we use unlabelled target 360 video data to train our model in a self-supervised manner. Empirical and qualitative results showed the potential of this work. We believe this work will inspire others to investigate this area of optical flow estimation.

\noindent \textbf{Acknowledgements:} This research was partially supported by NSF CSR-1908658 and NeTS-1909185. This article solely reflects the opinions and conclusions of its authors and not the funding agents.






%




\bibliographystyle{IEEEtran}
\bibliography{main}
\end{document}